\documentclass[10pt,journal,compsoc]{IEEEtran}
\ifCLASSOPTIONcompsoc
  \usepackage[nocompress]{cite}
\else
  \usepackage{cite}
\fi
\ifCLASSINFOpdf
\else
\fi
\usepackage{graphicx}
\usepackage{nameref,hyperref}

\usepackage{amsmath}
\begin{document}
%
\title{Cross-Cultural and Cultural-Specific Production and Perception of Facial Expressions of Emotion in the Wild}
%
%
%
%

\author{Ramprakash~Srinivasan,
        Aleix M.~Martinez
\IEEEcompsocitemizethanks{\IEEEcompsocthanksitem R. Srinivasan and A.M. Martinez are with the Department
of Electrical and Computer Engineering, The Ohio State University, Columbus, OH, 43210}
\thanks{Manuscript received August 01, 2018}}

%
%

\markboth{Journal of \LaTeX\ Class Files,~Vol.~14, No.~8, August~2015}%
{Shell \MakeLowercase{\textit{et al.}}: Bare Demo of IEEEtran.cls for Computer Society Journals}
%



\IEEEtitleabstractindextext{%
\begin{abstract}
Automatic recognition of emotion from facial expressions is an intense area of research, with a potentially long list of important application. Yet, the study of emotion requires knowing which facial expressions are used within and across cultures in the wild, not in controlled lab conditions; but such studies do not exist. Which and how many cross-cultural and cultural-specific facial expressions do people commonly use? And, what affect variables does each expression communicate to observers? If we are to design technology that understands the emotion of users, we need answers to these two fundamental questions. In this paper, we present the first large-scale study of the production and visual perception of facial expressions of emotion in the wild. We find that of the 16,384 possible facial configurations that people can theoretically produce, only 35 are successfully used to transmit emotive information across cultures, and only 8 within a smaller number of cultures. Crucially, we find that visual analysis of cross-cultural expressions yields consistent perception of emotion categories and valence, but not arousal. In contrast, visual analysis of cultural-specific expressions yields consistent perception of valence and arousal, but not of emotion categories. Additionally, we find that the number of expressions used to communicate each emotion is also different, e.g., 17 expressions transmit happiness, but only 1 is used to convey disgust.
\end{abstract}

}

\maketitle

\IEEEdisplaynontitleabstractindextext

%
\IEEEpeerreviewmaketitle

\IEEEraisesectionheading{\section{Introduction}\label{sec:introduction}}

%
%
%
%
\IEEEPARstart{A}{lthough} there is agreement that facial expressions are a primary means of social communication amongst people \cite{jack2017toward}, which facial configurations are {\em commonly} produced and successfully visually interpreted within and across cultures is a topic of intense debate \cite{barrett:18, barrett2017emotions, ekman2016scientists, kim2017human, martinez2017visual}. Resolving this impasse is essential to identify the validity of opposing models of emotion \cite{russell2017science}, and design technology that can interpret them \cite{martinez2017computational}.

Crucially, this impasse can only be solved if we study the production and perception of facial expressions in multiple cultures in the wild, i.e., outside the lab. This is the approach we take in this paper. Specifically, \textit{we identify commonly used expressions in the wild and assess their consistency of perceptual interpretation within and across cultures}. We present a computational approach that allows us to successfully do this on millions of images and billions of video frames.

\subsection{Problem definition}

Facial expressions are facial movements that convey emotion or other social signals robustly within a single or across multiple cultures \cite{ekman1992argument,russell2017science,izard2013human}. These facial articulations are produced by contracting and relaxing facial muscles, called Action Units (AUs) \cite{ekman1997face}. Typically, people employ 14 distinct AUs to produce facial configurations \cite{du2014compound}. Note we only consider the AUs that define facial expressions of emotion. We do not consider the AUs that specify distinct types of eye blinks (6 AUs), those that specify head position (8 AUs) or eye position (4 AUs), and those that are not observed in the expression of emotion (4 AUs); see \cite{ekman1997face,du2014compound} for a detailed discussion of the use of these AUs in the expression of emotion. 

Assuming each AU can be articulated independently, people are able to produce as many as 16,384 facial configurations. But how many of these facial configurations are expressions that communicate emotion?

We provide, to our knowledge, the first comprehensive, large-scale exploratory study of facial configurations \textit{in the wild}. It is crucial to note that we can only afford a successful answer to the above question if we study facial configurations outside the laboratory, i.e., in the wild. Unfortunately, to date, research has mostly focused on in lab studies and the analysis of a small number of sample expressions \cite{martinez2017visual}. In contrast, herein, we analyze over {\bf 7 million images} of facial configurations and {\bf 10,000 hours} of video filmed in the wild. Using an combination of automatic and manual analyses, we assess the consistency of production and visual perception of these facial configurations within and across cultures.

\subsection{Summary of our results}

Our study identifies 35 facial configurations that are consistently used across cultures. We refer to these expressions as cross-cultural. Our studies also identify 8 facial configurations that are used in some, but not all, cultures. These are termed cultural-specific expressions. This surprising result suggests that the number of facial expressions of emotion used within and across cultures is very small. The cross-cultural expressions represent \textless 0.22\% of all possible facial configurations. Cultural-specific expressions are \textless 0.05\% of all possible configurations.

We also investigate whether the visual analysis of these facial expressions yields consistent interpretations within and across cultures. We find that cross-cultural expressions yield consistent visual readings of emotion category and valence but not arousal. In contrast, cultural-specific expressions yield consistent inference of valence and arousal but not of emotion category. 

Importantly, the number of expressions used to communicate each emotion category is different. For example, 17 express happiness, 5 express anger, and 1 expresses disgust. 

Additionally, the accuracy of the visual interpretation of these expressions varies from a maximum of 89\% to a minimum of 22\%. This finding is fundamental if we wish to design technology that infers emotion from expressions. Some expressions communicate affect quite reliably, others do not. This also means that computer vision algorithms may be useful in some applications (e.g., searching for pictures in a digital photo album and in the web), but not in others (e.g., in the courts).  

Furthermore, the unexpected results reported in the present paper cannot be fully explained by current models of emotion and suggests there are multiple mechanisms involved in the production and perception of facial expressions of emotion. Affective computing systems will need to be design that make use of all the mechanisms described in this paper.
 
\subsection{Paper outline}

Section \ref{Sec: Results} details the design of the experiments and results. Section \ref{Sec: Discussion} discusses the significance and implications of the results. We conclude in Section \ref{Sec: Conclusions}

\section{Experiments and Results}\label{Sec: Results}

We performed a number of experiments to study the cross-cultural and cultural-specific uses of facial configurations.

\subsection{7 million images of facial configurations in the wild} 

The internet now provides a mechanism to obtain a large number of images of facial configurations observed in multiple cultures. We downloaded images of facial configurations with the help of major web search engines in countries where English, Spanish, Mandarin Chinese, Farsi, and Russian are the primary language \ref{Table 1}. We selected these languages because they (broadly) correspond to distinct grammatical families \cite{rowe2015concise} and are well represented on the web \cite{techs2014usage}.

Since we wish to find facial expressions associated with the communication of emotion, we used all the words in the dictionary associated with affect as keywords in our search. Specifically, we used WordNet \cite{miller1995wordnet}, a lexicon of the English language defined as a hierarchy of word relationships, including: synonyms (i.e., words that have the same or nearly the same meaning), hyponyms (i.e., subordinate nouns or nouns of more specific meaning), troponymys (i.e., verbs with increasingly specific meaning), and entailments (i.e., deductions or implications that follow logically from or are implied by another meaning). These hierarchies are given by a graph structure, with words represented as nodes and word relationships with directed edges pointing from more general words to more specific ones. Undirected edges are used between synonyms since neither word is more general than the other.

\begin{table}[]
\centering
\tiny
\begin{tabular}{llllll}
          & Chinese   & English       & Farsi & Russian & Spanish            \\ \hline
Countries & China     & United States & Iran  & Russia  & Spain              \\
          & Taiwan    & Canada        &       &         & Mexico             \\
          & Singapore & Australia     &       &         & Argentina          \\
          &           & Great Britain &       &         & Chile              \\
          &           &               &       &         & Peru               \\
          &           &               &       &         & Venezuela          \\
          &           &               &       &         & Colombia           \\
          &           &               &       &         & Ecuador            \\
          &           &               &       &         & Guatemala          \\
          &           &               &       &         & Cuba               \\
          &           &               &       &         & El Salvador        \\
          &           &               &       &         & Bolivia            \\
          &           &               &       &         & Honduras           \\
          &           &               &       &         & Dominican Republic \\
          &           &               &       &         & Paraguay           \\
          &           &               &       &         & Uruguay            \\
          &           &               &       &         & Nicaragua          \\
          &           &               &       &         & Costa Rica         \\
          &           &               &       &         & Puerto Rico (US)   \\
          &           &               &       &         & Panama             \\
          &           &               &       &         & Equatorial Guinea  	        
\end{tabular}
\caption{\textbf{Table 1}. Our image search was done using web search engines in the countries listed in this table. The behavioral experiment in Amazon Mechanical Turk (AMT) was open to residents of these countries too. Only people in the countries listed in each language were able to participate in the AMT experiment of that language group. Participants also had to pass a language test to verify proficiency on that language.}
\label{Table 1}
\end{table}

This graph structure allows us to readily identify words that describe affect concepts. They are the nodes emanating from the node ``feeling" in WordNet's graph. This yields a total of 821 words. Example words in our set are affect, emotion, anger, choler, ire, fury, madness, irritation, frustration, creeps, love, timidity, adoration, loyalty, happiness, etc. These 821 words were translated into Spanish, Mandarin Chinese, Farsi, and Russian by professional translators.

The words in each language were used as keywords to mine the web using search engines in multiple countries. Specifically, we used English words in web search engines typically employed in the United States, Canada, Australia, New Zealand and Great Britain; Spanish words in Spain, Mexico, Argentina, Chile, Peru, Venezuela, Colombia, Ecuador, Guatemala, Cuba, Bolivia, Dominican Republic, Honduras, Paraguay, Uruguay, El Salvador, Nicaragua, Costa Rica, Puerto Rico, Panama and Equatorial Guinea; Mandarin Chinese words in China, Taiwan and Singapore; Russian words in Russia; and Farsi words in Iran, table \ref{Table 1}. 

The process described above yielded a set of about 7.2 million images of facial configurations representing a large number of cultures. These facial configurations were then AU coded by a computer vision algorithm \cite{benitez2016emotionet}. To verify that the AU annotations provided by the computer vision system are accurate, we manually verify that at least ten sample images per language group were correctly labeled in each of the identified facial configurations. 

\begin{table*}[]
\centering
\begin{tabular}{ccc|ccc|ccc}
ID & AUs & Examples & ID & AUs             & Examples & ID & AUs             & Examples \\ \hline
1  & 4   	 & \includegraphics[width=0.30in]{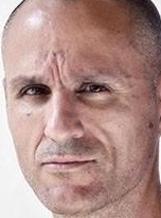}       & 13 & 4, 7, 9, 10, 17   & \includegraphics[width=0.30in]{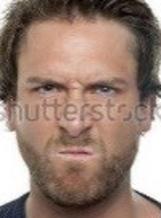}       & 25 & 1, 2, 5, 25, 26 		& \includegraphics[width=0.30in]{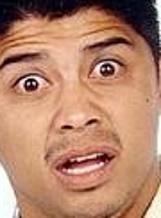}       \\
2  & 5           & \includegraphics[width=0.30in]{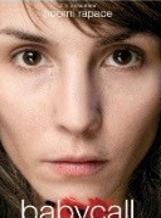}          & 14 & 9, 10, 15      & \includegraphics[width=0.30in]{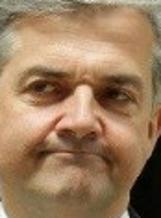}          & 26 & 4, 7, 9, 25, 26     & \includegraphics[width=0.30in]{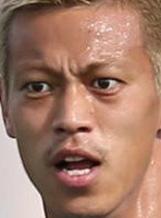}         \\
3  & 2, 4        & \includegraphics[width=0.30in]{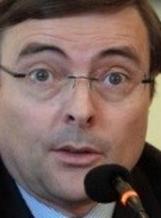}         & 15 & 12, 15          & \includegraphics[width=0.30in]{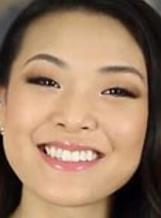}         & 27 & 12, 25, 26          & \includegraphics[width=0.30in]{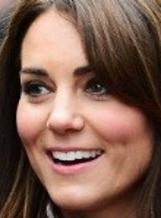}         \\
4  & 4, 7        & \includegraphics[width=0.30in]{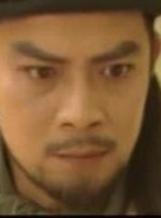}         & 16 & 6, 12, 25       & \includegraphics[width=0.30in]{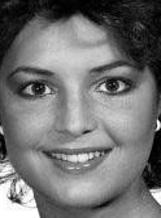}         & 28 & 2, 12, 25 26        & \includegraphics[width=0.30in]{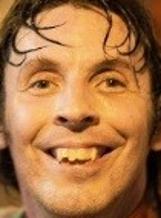}         \\
5  & 12          & \includegraphics[width=0.30in]{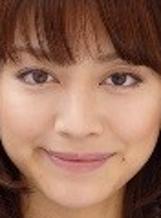}         & 17 & 2, 5, 12, 25    & \includegraphics[width=0.30in]{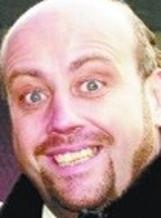}         & 29 & 5, 12, 25, 26       & \includegraphics[width=0.30in]{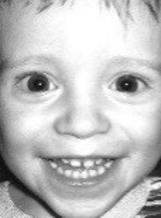}         \\
6  & 2, 12       & \includegraphics[width=0.30in]{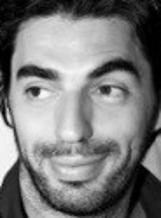}         & 18 & 1, 2, 5, 12, 25 & \includegraphics[width=0.30in]{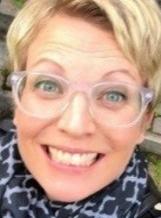}         & 30 & 2, 5, 12, 25, 26    & \includegraphics[width=0.30in]{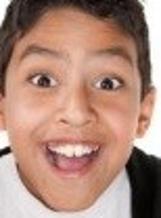}         \\
7  & 5, 12       & \includegraphics[width=0.30in]{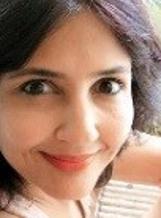}         & 19 & 6, 12, 25       & \includegraphics[width=0.30in]{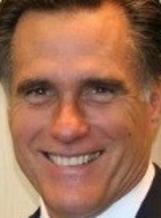}         & 31 & 1, 2, 5, 12, 25, 26 & \includegraphics[width=0.30in]{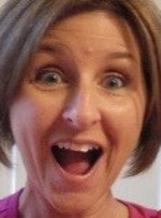}         \\
8  & 1, 2, 5, 12 & \includegraphics[width=0.30in]{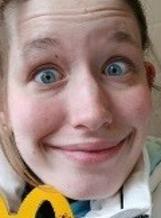}         & 20 & 10, 12, 25, 26  & \includegraphics[width=0.30in]{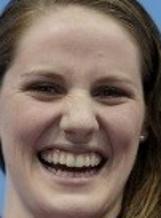}         & 32 & 6, 12, 25, 26       &  \includegraphics[width=0.30in]{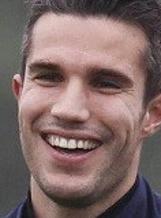}        \\
9  & 6, 12       & \includegraphics[width=0.30in]{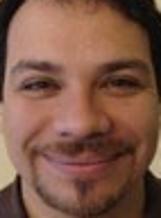}         & 21 & 1, 2, 25, 26    & \includegraphics[width=0.30in]{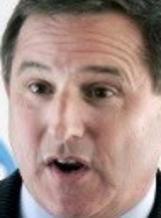}         & 33 & 4, 7, 9, 20, 25, 26 & \includegraphics[width=0.30in]{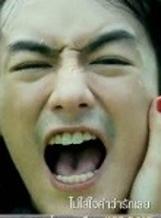}         \\
10 & 4, 15       & \includegraphics[width=0.30in]{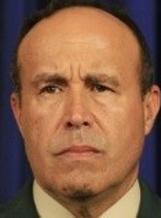}         & 22 & 1, 4, 25, 26   & \includegraphics[width=0.30in]{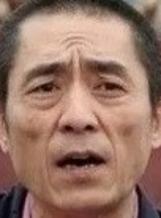}         & 34 & 1, 2, 5, 20, 25, 26 & \includegraphics[width=0.30in]{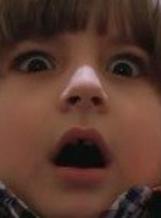}        \\
11 & 1, 4, 15    & \includegraphics[width=0.30in]{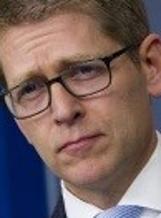}         & 23 & 5, 25, 26      & \includegraphics[width=0.30in]{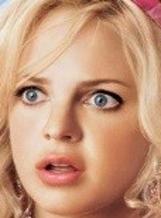}        & 35 & 1, 4, 5, 20, 26     &  \includegraphics[width=0.30in]{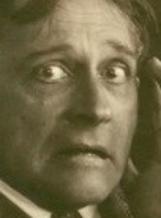}        \\
12 & 4, 7, 17    & \includegraphics[width=0.30in]{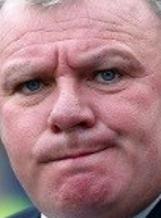}         & 24 & 2, 5, 25, 26   & \includegraphics[width=0.30in]{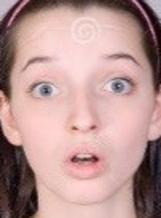}         &    &                     &        
\end{tabular}
\caption{Our study identified 35 unique combinations of AUs typically seen in the cultures where English, Spanish, Mandarin Chinese, Farsi, and Russian are the primary language. ID is the unique identification number given to each expression. AUs is the list of active action units in that expression. Also shown is a sample image of each expression.}\label{Fig: cross_cultural_faces}
\end{table*}

\subsection{Automatic recognition of facial configurations} 

Computer vision systems now provide reliable annotations of the most commonly employed AUs. We used the algorithm defined in \cite{benitez2016emotionet}, which provides accurate annotations of AUs in images collected in the wild. Specifically, we automatically annotated the 14 most common AUs in emotional expression (AUs 1, 2, 4, 5, 6, 7, 9, 10, 12, 15, 17, 20, 25 and 26) \cite{du2014compound}. To verify that the annotations provided by this computer vision algorithm were accurate, we manually verify that at least 50 images (10 per language group) in each possible facial configuration were correctly labeled. 

A {\em facial configuration} is defined by a unique set of active AUs. And, a facial expression is defined as a facial configuration that successfully transmits affect within and/or between cultures.

{\em Cross-cultural expressions}: For a facial configuration to be considered a possible cross-cultural expression of emotion, we require that at least $t$ sample images, in each of the cultural groups, be present in our database.

\begin{figure}[h]
\includegraphics[width=.48\textwidth]{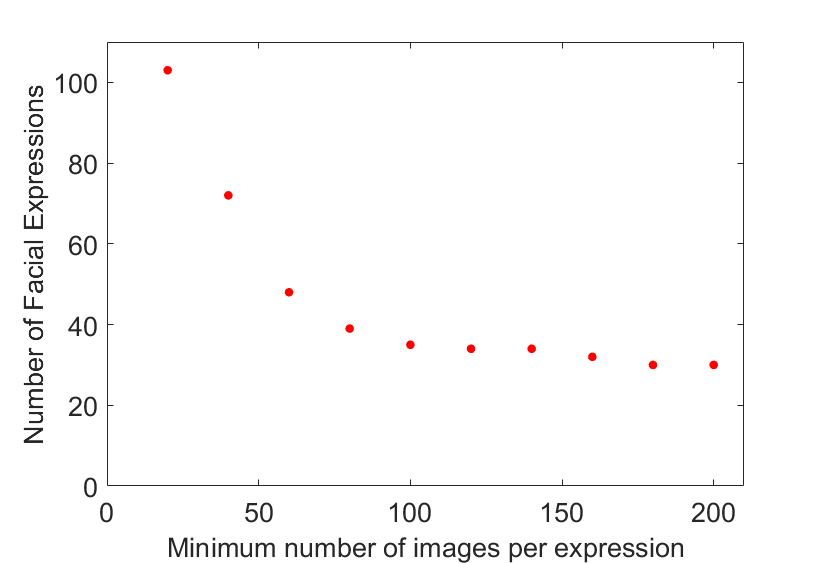}
\caption{Number of facial configurations in our dataset that have the number of images indicated in the x-axes.} \label{Fig: number_images_per_expn}
\end{figure}

The plot in Figure \ref{Fig: number_images_per_expn} shows the number of facial expressions identified with this approach in our database as a function of $t$. Note that the $x$-axis specifies the number of facial expressions, and the $y$-axis the value of $t$. 

We note that the number of expressions does not significantly change after $t=100$. Thus, in our experiments discussed below, we will set $t=100$.

{\em Cultural-specific expressions}: For a facial configuration to be considered a possible cultural-specific expression, we require that at least $t$ sample images in at least one cultural group are present in our database. When more than one cultural group, but not all, have $t$ or more samples of this facial configuration, this is also considered a cultural-specific expression.

\subsection{Number of cross-cultural facial expressions}
We wish to identify facial configurations that are commonly used across cultures. Thus, we identified the facial configurations with at least 100 image samples in each of the language groups. That is, for a facial configuration to be considered an expression, we required to have at least 100 sample images in each of the five language groups, for a total of 500 sample images. 

Using this criterion, we identified a total of 35 facial expressions. That is, we found 35 unique combinations of AUs with a significant number of sample images in each of the five language groups. No other facial configuration had a number of samples close to the required 500.

Table \ref{Fig: culture_specifc_faces} shows the list of active AUs and a sample image of each of these facial expressions. Each expression is given a unique ID number for future reference.

It is worth noting that 35 is a very small number; less than 0.22\% of the 16,384 possible facial configurations. This suggests that the number of facial configurations used to express emotion in the wild is very small. Nonetheless, the number of identified expressions is much larger than the classical six ``basic" emotions typically cited in the literature \cite{ekman1992argument}. In fact, these results are in line with newer computational and cognitive models of the production of facial expressions of emotion \cite{du2014compound, martinez2017visual}, which claim people regularly employ several dozen expressions. Our results are also consistent with the identification of at least 27 distinct experiences of emotion by \cite{cowen2017self}. 

An alternative explanation for our results is given by theoretical accounts of constructed emotions \cite{siegel2018emotion}. In this model, the identified expressions may be variations of some underlying dimensions of affect, rather than independent emotion categories. We assess this possibility in Section \ref{Sec: Visual Recognition}.

\begin{table*}[]
\begin{tabular}{cccc|ccccl}
ID & AUs          & Examples & Languages                  & ID & AUs                     & Examples & Languages                        \\ \hline
36 & 1            & \includegraphics[width=0.30in]{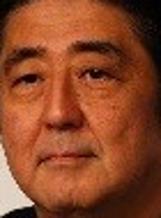}         & English, Mandarin, Spanish & 40 & 2, 5, 12                & \includegraphics[width=0.30in]{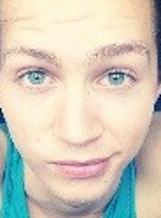}         & English, Russian, Farsi, Spanish \\
37 & 1, 2         & \includegraphics[width=0.30in]{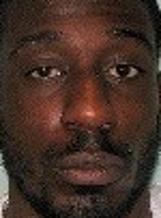}        & English, Russian, Spanish  & 41 & 4, 5, 10, 25, 26        & \includegraphics[width=0.30in]{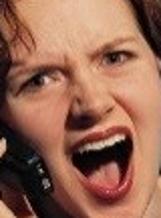}         & English, Farsi, Russian          \\
38 & 5, 17        & \includegraphics[width=0.30in]{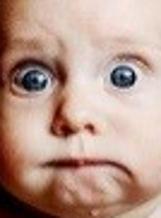}        & English, Mandarin          & 42 & 1, 25, 26               & \includegraphics[width=0.30in]{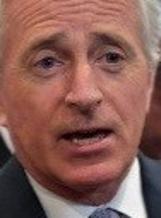}         & English                          \\
39 & 4, 7, 25, 26 & \includegraphics[width=0.30in]{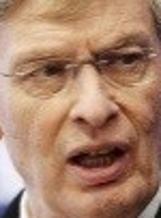}         & Farsi, Mandarin            & 43 & 4, 7, 9, 10, 20, 25, 26 & \includegraphics[width=0.30in]{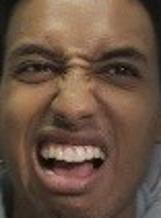}         & English                         
\end{tabular}
\caption{We identified only 8 combinations of AUs typically observed in some but not all cultures. ID is the unique identification number given to each expression. The ID numbers start at 36 to distinguish these facial expressions from those listed in Figure \ref{Fig: cross_cultural_faces}. The AUs column provides the list of active action units in that expression. Also shown is a sample image of each expression and a list of the language groups where each expression is commonly observed.}\label{Fig: culture_specifc_faces}
\end{table*}

\subsection{Number of cultural-specific facial expressions}

We ask whether the number of cultural-specific facial expressions is larger or smaller than that of cross-cultural expressions, as advocated by some models \cite{russell2017science}. To answer this question, we identify the number of facial configurations that have at least 100 sample images in one or more of the language groups, but not in all of them. 

This yielded a total of 8 expressions. Table \ref{Fig: culture_specifc_faces} shows a sample image and AU combination of each of these facial expressions.

Of the eight cultural-specific expressions, one expression was identified in four of the language groups, three expressions in three of the language groups, two expressions in two of the language groups, and two expressions in a single language group, Table \ref{Fig: culture_specifc_faces}. 

It is worth noting that of the eight cultural-specific expressions, seven include cultures were English is the primary language. Previous research suggests that Americans are more expressive with their faces than people in other cultures \cite{benitez2016not, rychlowska2015heterogeneity}. The results of the present paper are consistent with this hypothesis. 


Nonetheless, and unexpectedly, the number of cultural-specific expressions is smaller than that of expressions shared across language groups. Eight expressions correspond to less than 0.05\% of all possible facial configurations. And, of these, only two expressions were found in a single language. This is less than 0.013\% of all possible configurations. 

These results thus support the view that if a facial configuration is a facial expression of emotion, then this is much likely to be used in a large number of cultures than in a small number of them. This result could be interpreted as supporting the hypothesis that facial expressions primarily communicate emotion and other social signals across language groups \cite{darwin1998expression, duchenne1990mechanism, ekman1992argument, izard2013human}. But, for this hypothesis to hold, the visual interpretation of these facial expressions would also need to be consistent across language groups. We test this prediction in Section \ref{Sec: Visual Recognition}.

\subsection{Spontaneous facial expressions in video}

To verify that the 43 facial expressions identified above are also present in video of spontaneous facial expressions, we downloaded 10,000 hours of video from YouTube. This corresponds to over 1 billion frames. 

Specifically, we downloaded videos of documentaries, interviews, and debates where expressions are common and spontaneous. 

All video frames were automatically annotated by our computer vision system and verified manually; using the same approach described earlier in the paper. 

This analysis identified multiple instances of the 43 expressions of emotion listed in Tables \ref{Fig: cross_cultural_faces} and \ref{Fig: culture_specifc_faces}, further confirming the existence of these facial expressions of emotion in the wild.

\subsection{Visual recognition of cross-cultural facial expressions}\label{Sec: Visual Recognition}

\subsubsection{Are the facial expressions of emotion identified above visually recognized across cultures?} 

Some theorists propound that people should be able to infer emotion categories across cultures (e.g., happy, sad, and angrily surprised) \cite{darwin1998expression, du2014compound, ekman1992argument}, while others argue that expressions primarily communicate valence and arousal \cite{russell2017science, posner2005circumplex, russell2003core}.

\begin{figure}
\centering
\includegraphics[width=0.45\textwidth]{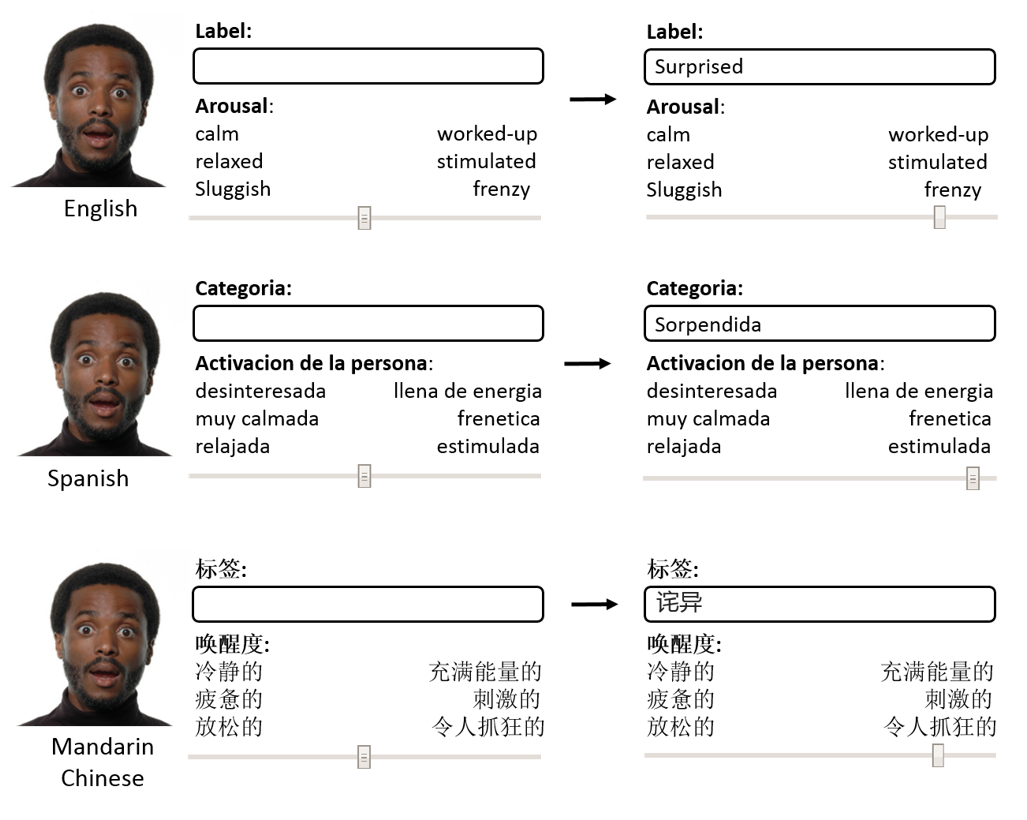}
\caption{Typical timeline of the behavioral experiment completed by participants in Amazon Mechanical Turk. Shown here is an example of a possible response given in three languages (English, Spanish and Mandarin Chinese).}\label{Fig: timeline_AMT}
\end{figure}

To test which of the above affect variables are consistently identified across language groups, we performed an online behavioral experiment. In this experiment, participants evaluated the 35 expressions observed in all language groups. Here, 50 images of each of the 35 expressions were shown to participants, one at a time, Figure \ref{Fig: timeline_AMT}. 

Participants were asked to provide an unconstrained label (1 to 3 words) describing the emotion conveyed by the facial expression and to indicate the level of arousal on a 6-point scale. Valence was obtained by identifying whether the word (or words) given by the participant was (were) of positive, neutral or negative affect as given in a dictionary. Details of the methods of this behavioral experiment and its computational analysis are in Section \ref{behavioral_expt}.

To deal with the many words one may use to define the same emotion, responses were aggregated to the most general word in WordNet. That is, if a subject selected a word that is a synonym or a descendant of a node used by another subject, the former was counted as a selection of the latter. For instance, if a subject used the word ``anger" and another the word ``irritated" to label the same facial expression, then the two were counted as a selection of ``anger" because ``anger" is an ascendant of ``irritated" in WordNet.

A total of 150 subjects participated in this experiment, 30 subjects per language group. Each participant evaluated the 1,750 images in the experiment. Subjects in each language group had to be from a country where that language is spoken, Table \ref{Table 1}. Participants provided answers in their own language. Non-English responses were translated into English by an online dictionary.

\subsubsection{Is there consistency of emotion categorization across language groups?} 

We find that the most commonly selected word was indeed identical in all language groups, except for expression ID 3 (which is produced with AUs 2 and 5), Table \ref{Table: common_label}. 

\begin{figure*}
\includegraphics[width=.97\textwidth]{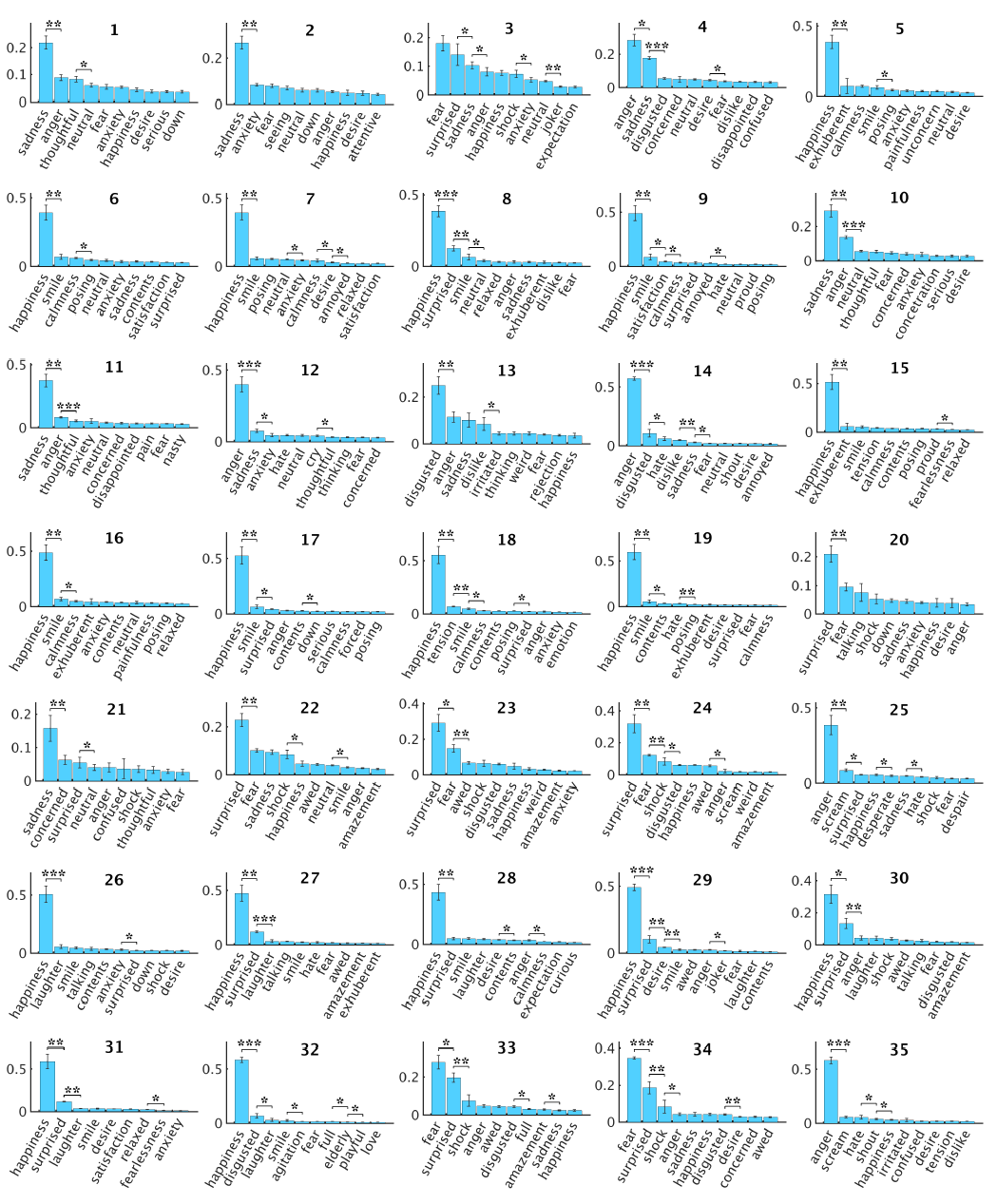}
\caption{Categorical labeling of the expressions observed across cultures. The y-axis indicates the proportion of times subjects in the five language groups selected the emotion category labels given in the x-axis. The center number in each plot correspond to the ID numbers in Figure \ref{Fig: cross_cultural_faces}. * p \textless .01, ** p \textless .001, *** p \textless .001.}\label{Fig: cross_categories}
\end{figure*}

We also computed statistical significance between word selections. Here too, we find that the difference between the top selection and others are statistically significant except (again) for expression ID 3, Figure \ref{Fig: cross_categories}. 

As can be appreciated in the figure, however, some labels were selected more consistently than others. For example, the first expression (produced with AU 4) is perceived as expressing \textit{sadness} (the top selected word) about 22\% of the time, but the last expression (produced with AUs 1, 4, 5, 20, 25 and 26) is perceived as expressing \textit{anger} about 60\% of the time. 

\begin{table}[]
\centering
\begin{tabular}{llllll}
\textbf{ID} & \textbf{Chinese}   & \textbf{English}   & \textbf{Farsi}     & \textbf{Russian}   & \textbf{Spanish}   \\ \hline
1           & \textbf{sadness}   & \textbf{sadness}   & \textbf{sadness}   & \textbf{sadness}   & \textbf{sadness}   \\
2           & \textbf{sadness}   & \textbf{sadness}   & \textbf{sadness}   & \textbf{sadness}   & \textbf{sadness}   \\
3           & fear               & sadness            & fear               & surprise           & surprise           \\
4           & \textbf{anger}     & \textbf{anger}     & \textbf{anger}     & \textbf{anger}     & \textbf{anger}     \\
5           & \textbf{happiness} & \textbf{happiness} & \textbf{happiness} & \textbf{happiness} & \textbf{happiness} \\
6           & \textbf{happiness} & \textbf{happiness} & \textbf{happiness} & \textbf{happiness} & \textbf{happiness} \\
7           & \textbf{happiness} & \textbf{happiness} & \textbf{happiness} & \textbf{happiness} & \textbf{happiness} \\
8           & \textbf{happiness} & \textbf{happiness} & \textbf{happiness} & \textbf{happiness} & \textbf{happiness} \\
9           & \textbf{happiness} & \textbf{happiness} & \textbf{happiness} & \textbf{happiness} & \textbf{happiness} \\
10          & \textbf{sadness}   & \textbf{sadness}   & \textbf{sadness}   & \textbf{sadness}   & \textbf{sadness}   \\
11          & \textbf{sadness}   & \textbf{sadness}   & \textbf{sadness}   & \textbf{sadness}   & \textbf{sadness}   \\
12          & \textbf{anger}     & \textbf{anger}     & \textbf{anger}     & \textbf{anger}     & \textbf{anger}     \\
13          & \textbf{disgust}   & \textbf{disgust}   & \textbf{disgust}   & \textbf{disgust}   & \textbf{disgust}   \\
14          & \textbf{anger}     & \textbf{anger}     & \textbf{anger}     & \textbf{anger}     & \textbf{anger}     \\
15          & \textbf{happiness} & \textbf{happiness} & \textbf{happiness} & \textbf{happiness} & \textbf{happiness} \\
16          & \textbf{happiness} & \textbf{happiness} & \textbf{happiness} & \textbf{happiness} & \textbf{happiness} \\
17          & \textbf{happiness} & \textbf{happiness} & \textbf{happiness} & \textbf{happiness} & \textbf{happiness} \\
18          & \textbf{happiness} & \textbf{happiness} & \textbf{happiness} & \textbf{happiness} & \textbf{happiness} \\
19          & \textbf{happiness} & \textbf{happiness} & \textbf{happiness} & \textbf{happiness} & \textbf{happiness} \\
20          & \textbf{surprise}  & \textbf{surprise}  & \textbf{surprise}  & \textbf{surprise}  & \textbf{surprise}  \\
21          & \textbf{sadness}   & \textbf{sadness}   & \textbf{sadness}   & \textbf{sadness}   & \textbf{sadness}   \\
22          & \textbf{surprise}  & \textbf{surprise}  & \textbf{surprise}  & \textbf{surprise}  & \textbf{surprise}  \\
23          & \textbf{surprise}  & \textbf{surprise}  & \textbf{surprise}  & \textbf{surprise}  & \textbf{surprise}  \\
24          & \textbf{surprise}  & \textbf{surprise}  & \textbf{surprise}  & \textbf{surprise}  & \textbf{surprise}  \\
25          & \textbf{anger}     & \textbf{anger}     & \textbf{anger}     & \textbf{anger}     & \textbf{anger}     \\
26          & \textbf{happiness} & \textbf{happiness} & \textbf{happiness} & \textbf{happiness} & \textbf{happiness} \\
27          & \textbf{happiness} & \textbf{happiness} & \textbf{happiness} & \textbf{happiness} & \textbf{happiness} \\
28          & \textbf{happiness} & \textbf{happiness} & \textbf{happiness} & \textbf{happiness} & \textbf{happiness} \\
29          & \textbf{happiness} & \textbf{happiness} & \textbf{happiness} & \textbf{happiness} & \textbf{happiness} \\
30          & \textbf{happiness} & \textbf{happiness} & \textbf{happiness} & \textbf{happiness} & \textbf{happiness} \\
31          & \textbf{happiness} & \textbf{happiness} & \textbf{happiness} & \textbf{happiness} & \textbf{happiness} \\
32          & \textbf{happiness} & \textbf{happiness} & \textbf{happiness} & \textbf{happiness} & \textbf{happiness} \\
33          & \textbf{fear}      & \textbf{fear}      & \textbf{fear}      & \textbf{fear}      & \textbf{fear}      \\
34          & \textbf{fear}      & \textbf{fear}      & \textbf{fear}      & \textbf{fear}      & \textbf{fear}      \\
35          & \textbf{anger}     & \textbf{anger}     & \textbf{anger}     & \textbf{anger}     & \textbf{anger}     \\ \hline
\end{tabular}
\caption{Most commonly used emotion category labels by subjects of the five language groups.} \label{Table: common_label}
\end{table}

Thus, although there is a cross-cultural agreement on the categorical perception of the herein-identified facial expressions, this consensus varies wildly from one expression to the next. 

Most current models and affect computing algorithms of the categorical perception of facial expressions of emotion do not account this large variability.  These results are an urgent call to researchers to start coding this fundamental information in their algroithms. Otherwise, the outcome of emotion inference will carry little value, because it is not always a robust signal. 

Crucially too, we note that several facial configurations are categorized as expressing the same emotion, Table \ref{Table: common_label}. For example, anger and sadness are expressed using five distinct facial configurations each, whereas disgust is conveyed using a single facial configuration. This variability in number of facial configurations per emotion category is not accounted for by current models of the production of facial expressions, but it is essential to understand emotion.

\begin{figure*}
\includegraphics[width=.97\textwidth]{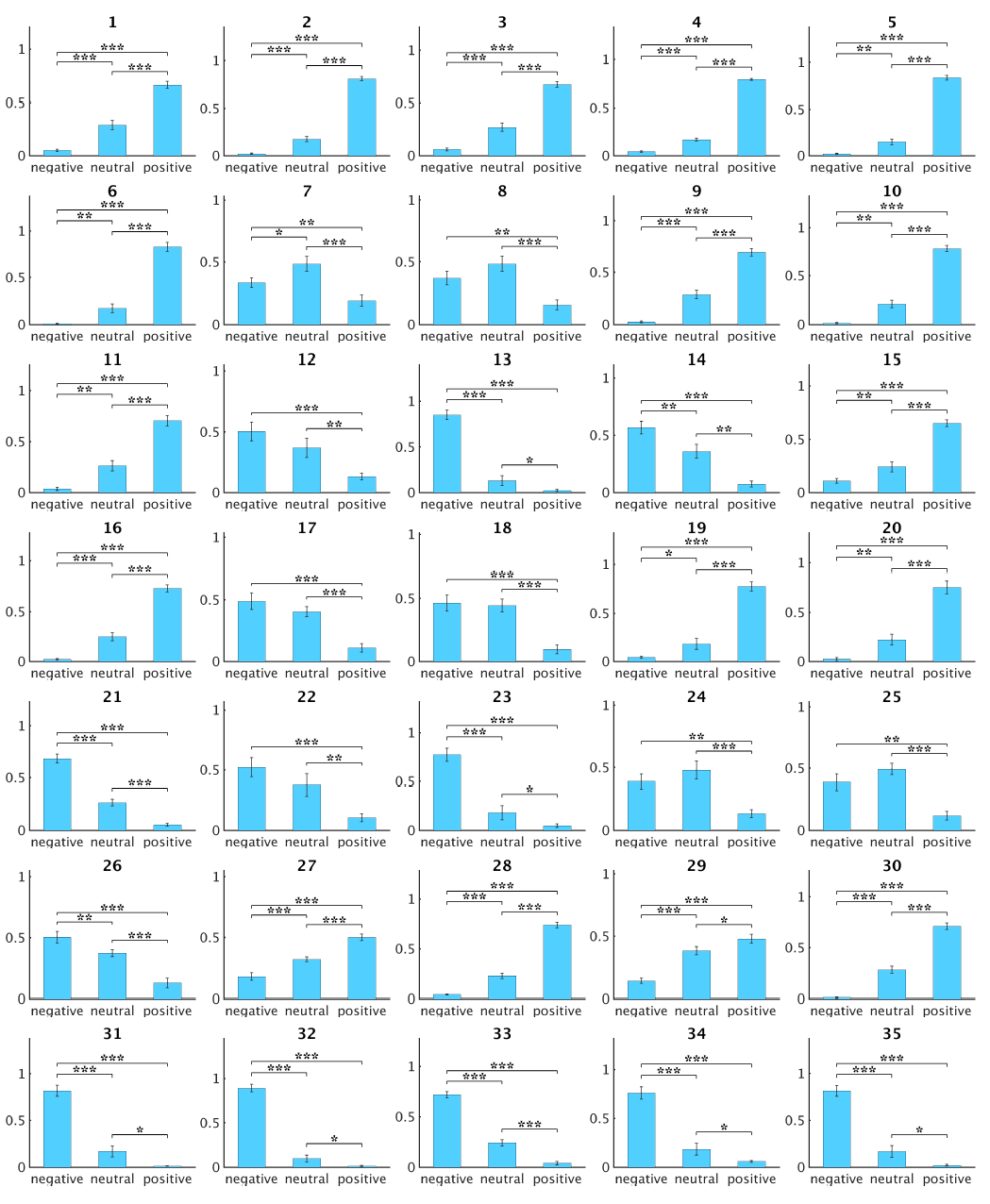}
\caption{Perception of valence of the facial expression used across cultures. The y-axis indicates the proportion of times subjects in the five language groups selected positive, neutral and negative labels to describe each facial configuration. Thus, the x-axis specifies valence. The top-center number in each plot correspond to the ID of the facial configuration as given in Figure \ref{Fig: cross_cultural_faces}. * p \textless .01, ** p \textless .001, *** p \textless .001.} \label{Fig: Cross_valence}
\end{figure*}

Next, we ask if the consistency in the perception of valence is stronger than that of emotion categorization. Figure \ref{Fig: Cross_valence} shows the percentage of times subjects selected a negative, neutral and positive word to define each expression. Statistical differences between negative, neutral and positive valence are also specified in the figure. As we can see in the figure, there is overall strong agreement amongst participants of different cultures on the valence of the expression. 

These results are in favor of a cross-cultural perception of valence in each of the 35 facial expressions, including the expression ID 3 for which emotion categorization was not consistent. Nonetheless, as with emotion categorization, the degree of agreement across expressions varies significantly, from a maximum of 89.05\% to a minimum of 46.25\%. This variability in effect size is unaccounted for in current models and algroithms. One possible way to address this problem is to include the modeling of personal experiences and context \cite{barrett2017emotions, leitzke2016developmental}, but it is currently unknown how this modeling would yield the results reported in this paper. We urge researchers to address this open problem.

\begin{figure*}[h!]
{\small\bf a.}\includegraphics[width=.48\textwidth]{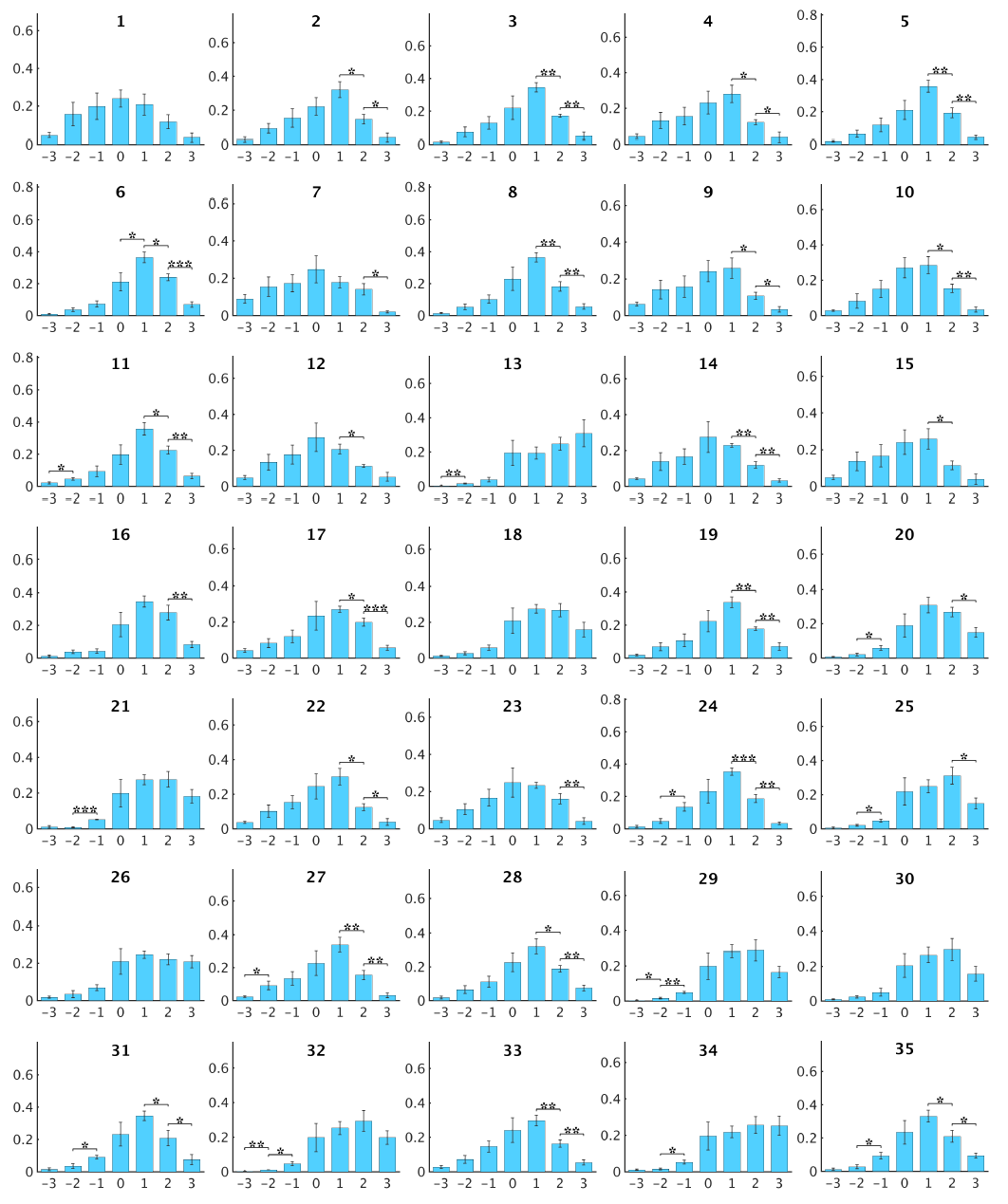}
{\small\bf b.}\includegraphics[width=.48\textwidth]{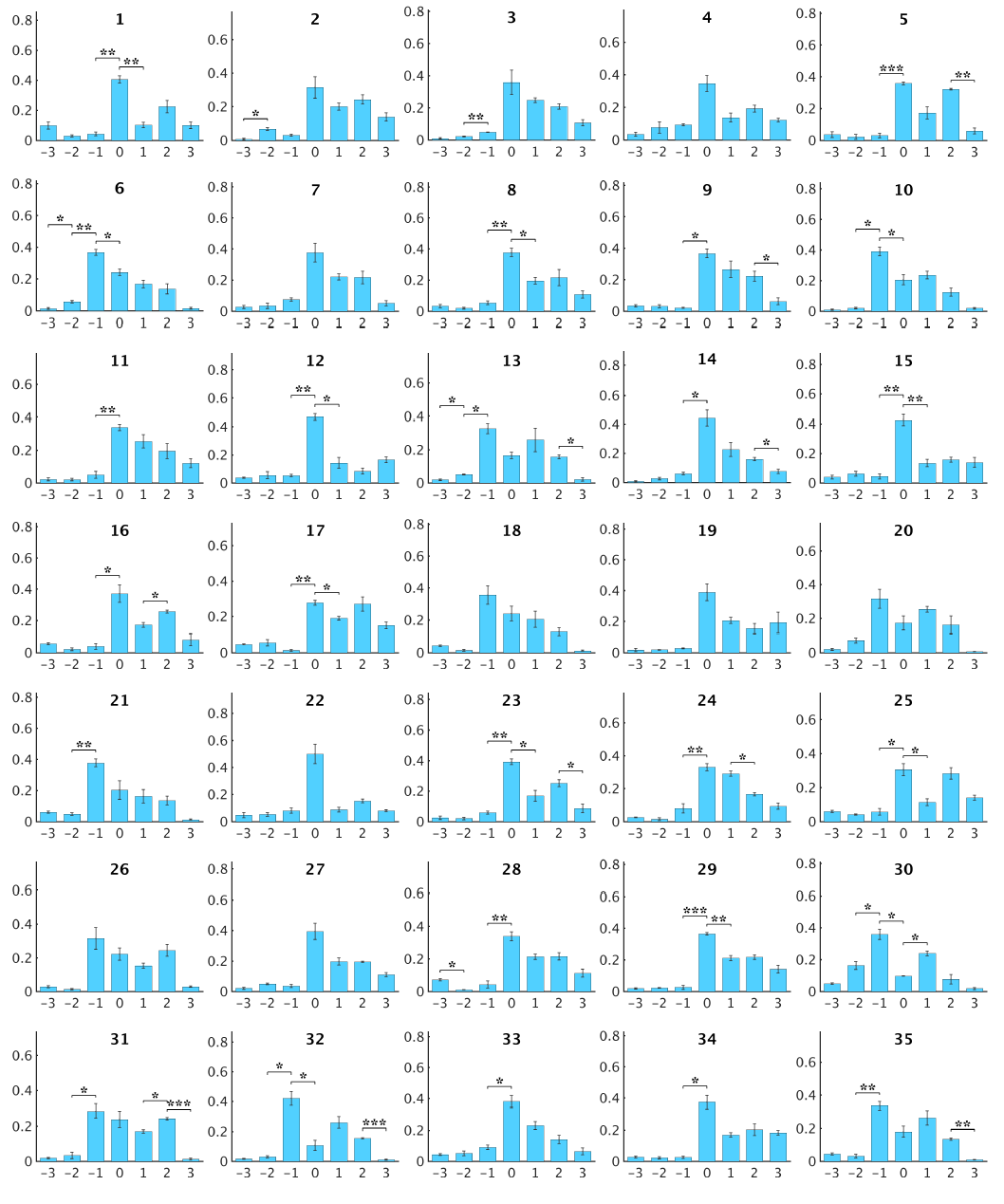}
{\small\bf c.}  \includegraphics[width=.48\textwidth]{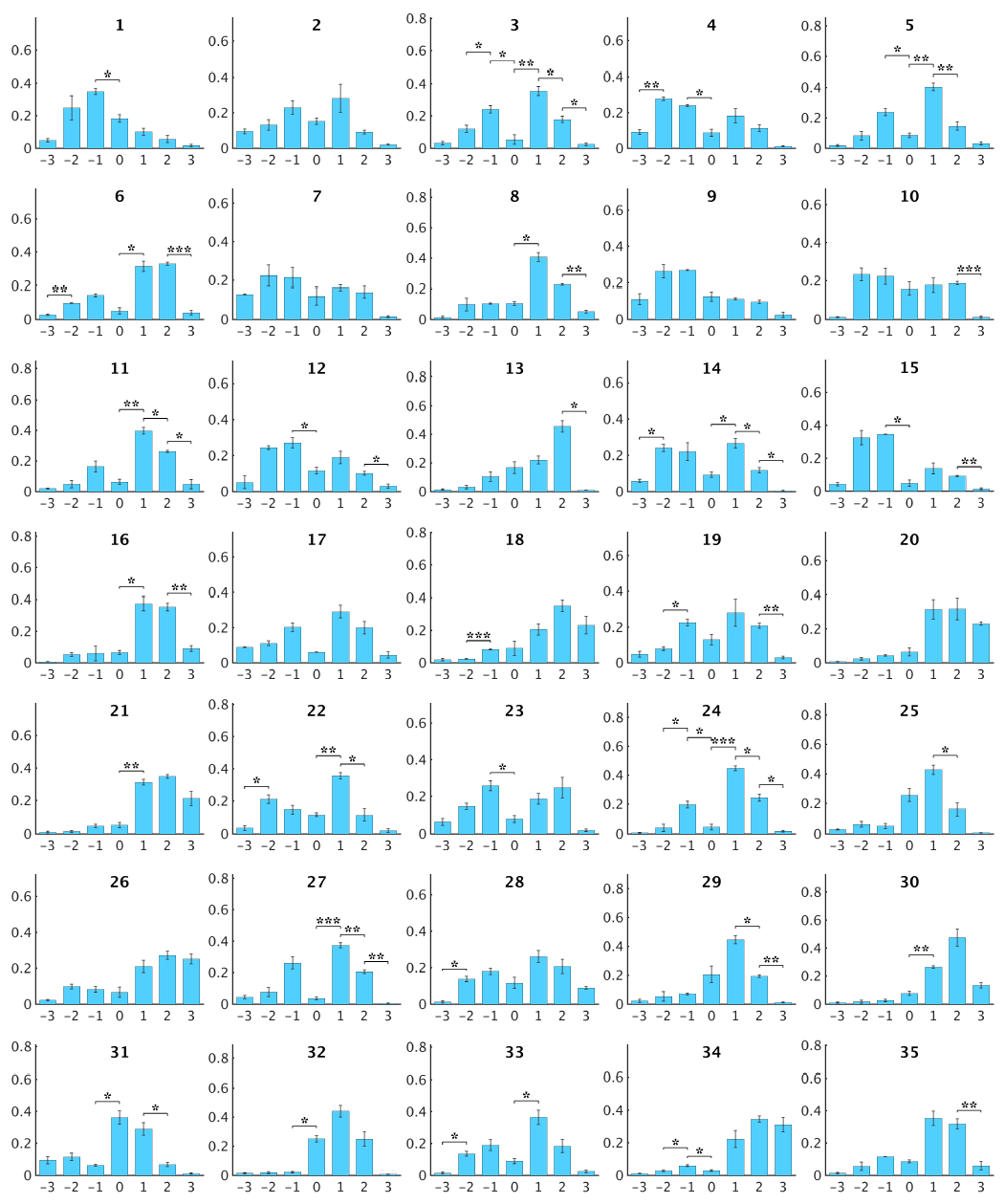}
{\small\bf d.} \includegraphics[width=.48\textwidth]{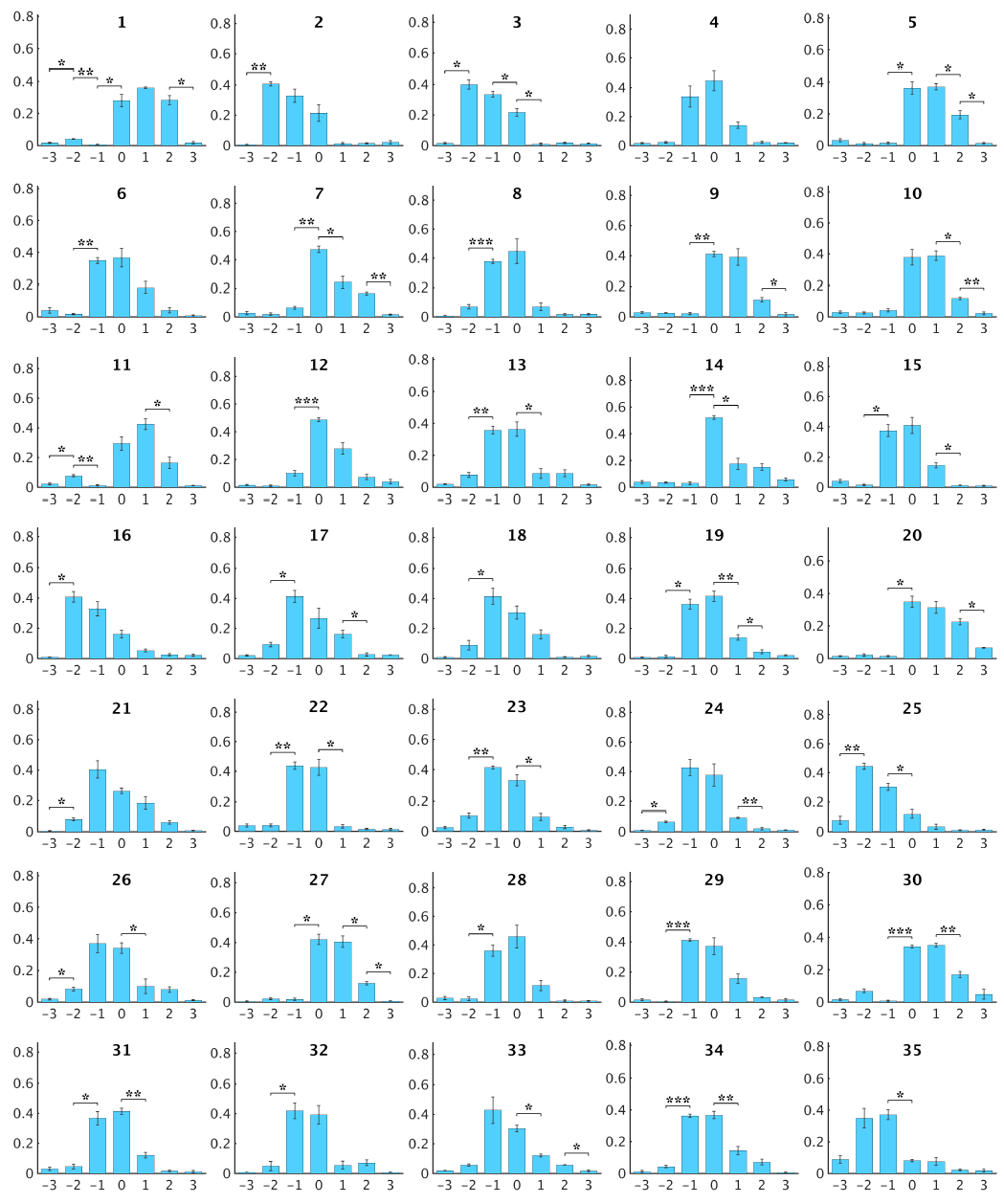}
\caption{{\bf a.} Perception of arousal of the facial expression used across cultures. The plots show the proportion of subjects'€™ responses ($y$-axis) for each level of arousal (x-axis).  {\bf b.} Perception of arousal of the cross-cultural expressions by subjects in the Mandarin Chinese language group. The $y$-axis indicates the proportion of times subjects selected each of the levels of arousal indicated in the $x$-axis. {\bf c.} Perception of arousal of the cross-cultural expressions by subjects in the English language group. {\bf d.} Perception of arousal of the cross-cultural expressions by subjects in the Farsi language group.  top, center number in each plot correspond to the ID numbers in Figure \ref{Fig: cross_cultural_faces}.  * p \textless .01, ** p \textless .001, *** p \textless .001.}  \label{Fig: Cross_arousal}\end{figure*}

\begin{figure*}[h]
{\small\bf a.}\includegraphics[width=.48\textwidth]{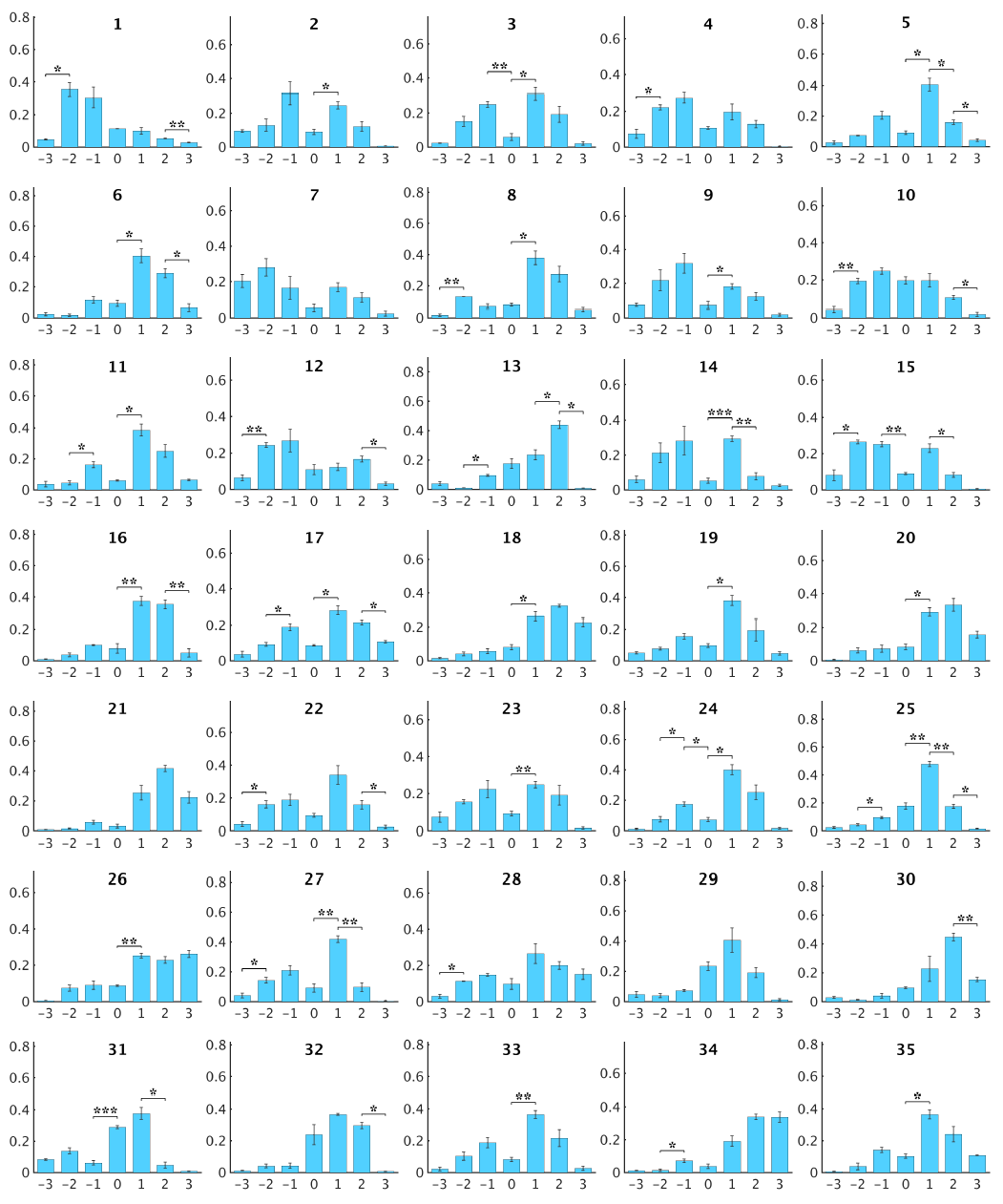}
{\small\bf b.}\includegraphics[width=.48\textwidth]{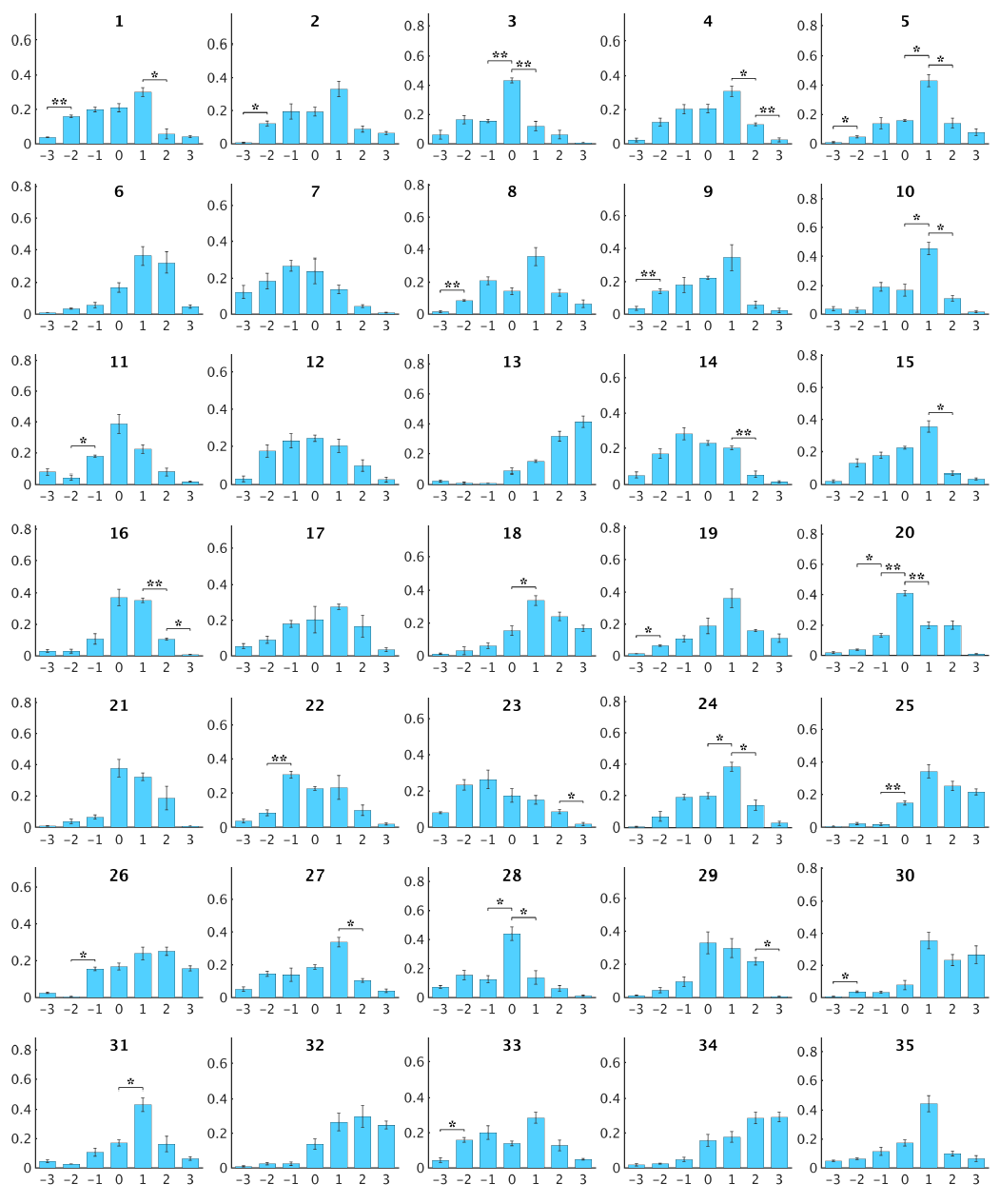}
\caption{{\bf a.} Perception of arousal of the cross-cultural expressions by subjects in the Spanish language group.. {\bf b.} Perception of arousal of the cross-cultural expressions by subjects in the Russian language group. The $y$-axis indicates the proportion of times subjects selected each of the levels of arousal indicated in the $x$-axis. The top, center number in each plot correspond to the ID numbers in Figure \ref{Fig: cross_cultural_faces}. * p \textless .01, ** p \textless .001, *** p \textless .001.} \label{Fig: arousal_spanish}  
\end{figure*}

Since we observed cross-cultural consistency in the perception of emotion category and valence, it is reasonable to assume the same will be true for arousal. However, we find that the selection of arousal yields little to no agreement amongst subjects of different or even the same language groups, Figure \ref{Fig: Cross_arousal}{\bf a} and Figures \ref{Fig: Cross_arousal}{\bf b-d} and \ref{Fig: arousal_spanish}{\bf a-b}.

Additionally, the effect size of the top selection of arousal is generally very low. Also, as it is appreciated in the figure, there are very few selections that are statistically different from the rest. This suggests a mostly random selection process by human observers. There is a mild tendency toward agreement in the facial expressions with IDs 6 and 24. But, even in these cases, the results are too weak to suggest successful transmission of arousal. 

\subsubsection{Methods of the behavioral experiment}\label{behavioral_expt} 

Once the set of facial expressions are identified, we performed a behavioral experiments.

Our participants had normal or corrected-to-normal vision. We assessed consistency of perception of the above-identified facial expressions by native subjects living in the countries listed in Table \ref{Table 1}. This was an online experiment performed in Amazon Mechanical Turk. Country of residence was checked by IP address, and proficiency in the major language of that country using a short grammatical test. A total of 150 subjects participated in this experiment, 30 subjects per language group. Before the experiment started, participants watched a short video which provided instructions on how to complete the task. The instructions and video were in the language of the participant. A small monetary payment was provided for completion of the experiment.

Participants assessed a total of 1,750 images. Images were displayed one at a time, Figure \ref{Fig: timeline_AMT}. A textbox was provided right of the image. Participants were asked to enter a 1 to 3-word description of the emotion expressed in the facial expression shown on left, Figure \ref{Fig: timeline_AMT}. Below this textbox, participants had a sliding bar. This bar was used to indicate the level of arousal perceived in the expression on left. The bar could be moved into six positions, going from disengaged (calm, relax, sluggish) to engaged (worked-up, stimulated, frenzy). The experiment was self-paced, i.e., there was no limit on time to complete this task. 

Category labels provided by participants in a language other than English were automatically translated into English using the tools given in the Natural Language Toolkit (NLTK) \cite{bird2004nltk}. The labels of all languages were then concatenated. This yielded the number of times each label was used by participants. 

Next, we reduced the number of word redundancies. This was done using the hierarchical structure provided in WordNet. Specifically, any label that was a descendant of another label was converted to the latter, e.g., if a subject labeled a facial expression as expressing ``outrage" and another subject labeled the same face as expressing ``anger", then ``outraged"" would be converted into ``anger" because anger is an ancestor of outraged in WordNet. 

Also, some participants provided non-emotive labels. For example, ``looks angry" was a typical response. To eliminate labels that do not define an emotion concept (e.g.,``looks"), we eliminated words that are not a descendant of the word ``feeling" in WordNet. 

This process yielded the final word tally for each facial configuration. Frequency of word selection was given as the fraction of times each word was selected over the total number of words chosen by participants. Standard deviations of these frequencies were also computed. A paired sample right-tailed $t$-test was performed to test whether each word is significantly more frequent than the subsequent word (i.e., most selected word versus second most selected word, second most selected word versus third most selected word, etc.). 

Next, we used a dictionary to check the valence (positive, neutral, or negative) of each of the words obtained by the above procedure. This was given automatically by the semantic processing tools in NLTK, which identifies the valence of a word. This allowed us to compute the number of times participants selected positive, neutral and negative valence in each facial configuration. The same frequency and standard deviation computations described above are reported. Here, we used double-tailed $t$-tests to determine statistical difference between positive versus neutral valence, positive versus negative valence, and neutral versus negative valence.

Finally, the same procedure described above was used to determine the frequency of selection of each arousal level as well as their standard deviation. A double-tailed $t$-test was used to test statistical significance between neighboring levels of arousal.

\subsection{Visual recognition of cultural-specific facial expressions}

We ask if the results obtained above replicate when we study facial expressions that are only common in some (but not all) of the five language groups. As in the above, we first assessed the consistency in categorical labeling of the eight cultural-specific expressions. For each expression, only subjects in the language groups where this expression was found to be commonly used were asked to provide categorical labels. 

\begin{figure}[h]
\includegraphics[width=.48\textwidth]{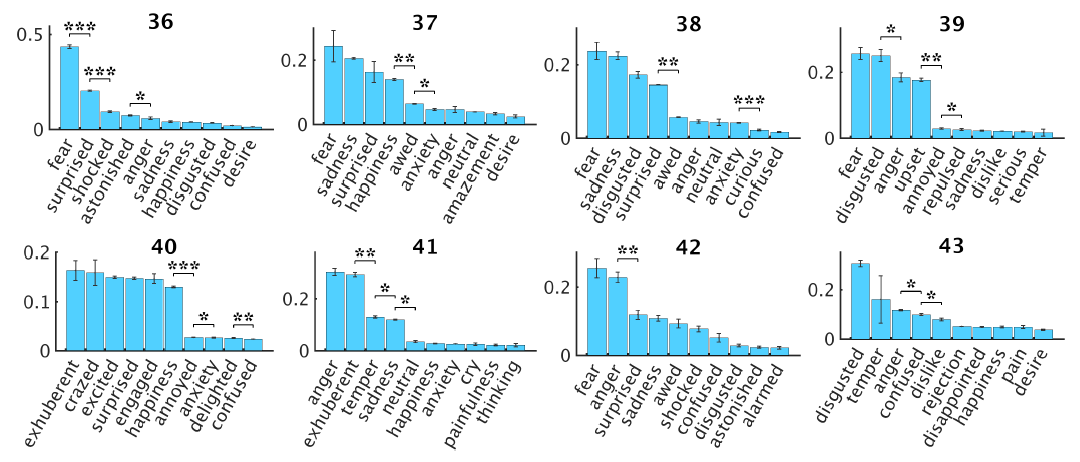}
\caption{The y-axis indicates the proportion of times subjects in the language groups where that expression is typically observed selected each of emotion category labels given in the x-axis. The center number in each plot correspond to the ID numbers in Figure \ref{Fig: culture_specifc_faces}.  * p \textless .01, ** p \textless .001, *** p \textless .001.}\label{Fig: specific_categories}
\end{figure}

As we can see in Figure \ref{Fig: specific_categories}, there is very little agreement in labeling these cultural-specific expressions, except for expression ID 36. This is in sharp contrast with the labeling results of the cross-cultural expressions shown in Figure \ref{Fig: cross_categories}. 

These results suggest that cross-cultural expressions are employed in multiple cultures precisely because they provide a relatively accurate communication of emotion category. But, cultural-specific expressions do not. 

\begin{figure}[h]
\includegraphics[width=.48\textwidth]{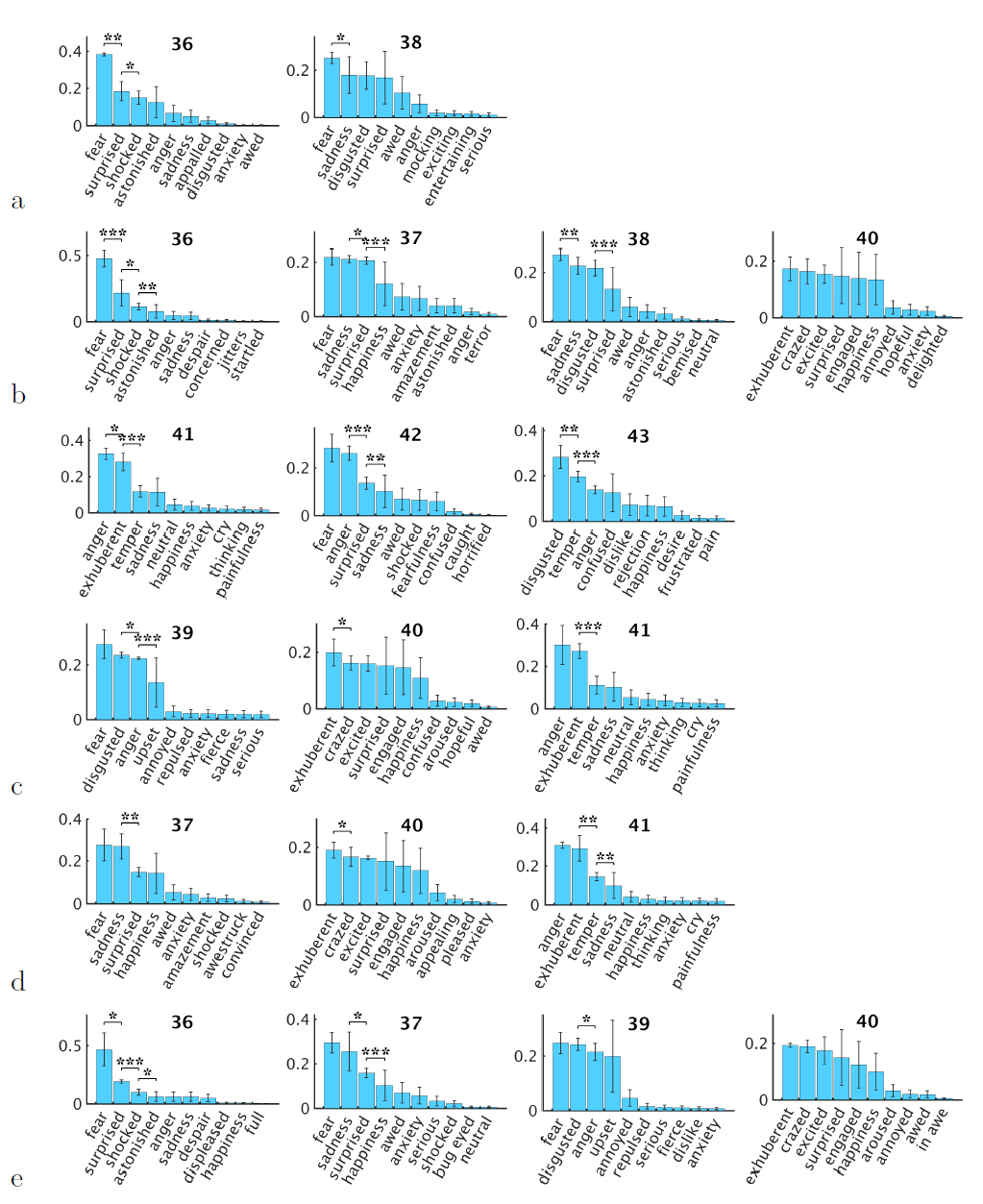}
\caption{Categorical labeling of the cultural-specific expressions by subjects in each of the language groups: Mandarin Chinese (a), English (b), Farsi (c), Russian (d), and Spanish (e). The y-axis indicates the proportion of times subjects selected the emotion category label given in the x-axis. The top, center number in each plot correspond to the ID numbers in Figure \ref{Fig: culture_specifc_faces}. * p \textless .01, ** p \textless .001, *** p \textless .001.} \label{Fig: specific_categories_individual}
\end{figure}

What is the purpose of cultural-specific expressions then? Do we see consistency within each individual culture? As can be seen in Figure \ref{Fig: specific_categories_individual}, the answer is again no. There is little agreement in labeling cultural-specific expressions even within the same culture.

Although cultural-specific expressions do not yield a consistent perception of emotion category, maybe they transmit some other affect information. 

\begin{figure}
\includegraphics[width=.48\textwidth]{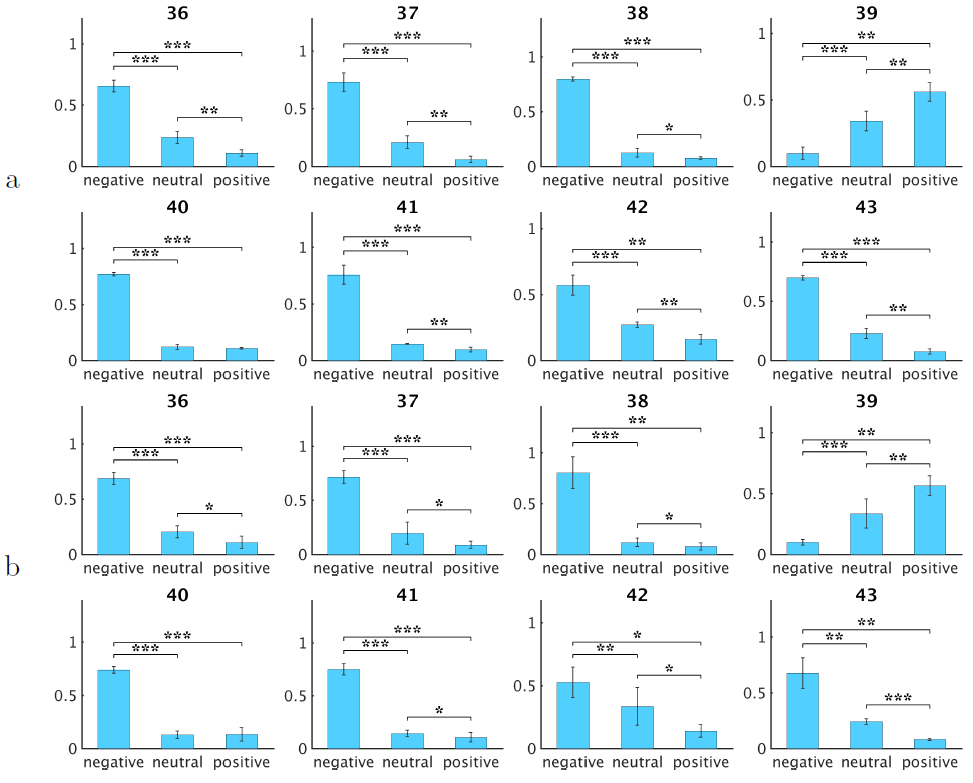}
\caption{\textbf{a.} Perception of valence of the cultural-specific expression by subjects in the cultures where this expression is common. The plots show the proportion of times (y-axis) subjects selected positive, neutral and negative labels (x-axis) to describe each facial configuration. \textbf{b.} Perception of valence of the eight cultural-specific expressions by subjects in the cultures where the expressions are not commonly observed. The top-center number in each plot correspond to the ID of the facial configuration as given in Figure \ref{Fig: culture_specifc_faces}. * p \textless .01, ** p \textless .001, *** p \textless .001.} \label{Fig: specific_valence}
\end{figure}

As in the above, we ask if valence is consistently perceived in the cultures where these cultural-specific expressions are observed? Figure \ref{Fig: specific_valence}a summarizes the results. 

It is clear from these results that valence is robustly and consistently interpreted by subjects of all cultures where the expression is used. In fact, the agreement across participants is generally very high. 

This surprising finding, which contrasts with that of emotion categorization, suggest that {\em while cross-cultural expressions may be used to transmit emotion category and valence, cultural-specific expression may only transmit valence. }

Given the above result, we wondered if the perception of valence in cultural-specific expressions is also consistently interpreted in cultures where these expressions are not commonly observed. 

Figure \ref{Fig: specific_valence}b shows the perception of valence of the eight cultural-specific expressions by subjects in cultures where the expressions are not commonly seen. 

As can be seen in this figure, subjects in these cultures successfully interpret valence as well. These results support the view that the valence of cultural-specific expressions is consistently transmitted across cultures, even if those expressions are not typically seen in them.

Since the role of cultural-specific expressions seems to be that of communicating valence, we ask if they can also successfully transmit arousal. 

\begin{figure}
\includegraphics[width=.48\textwidth]{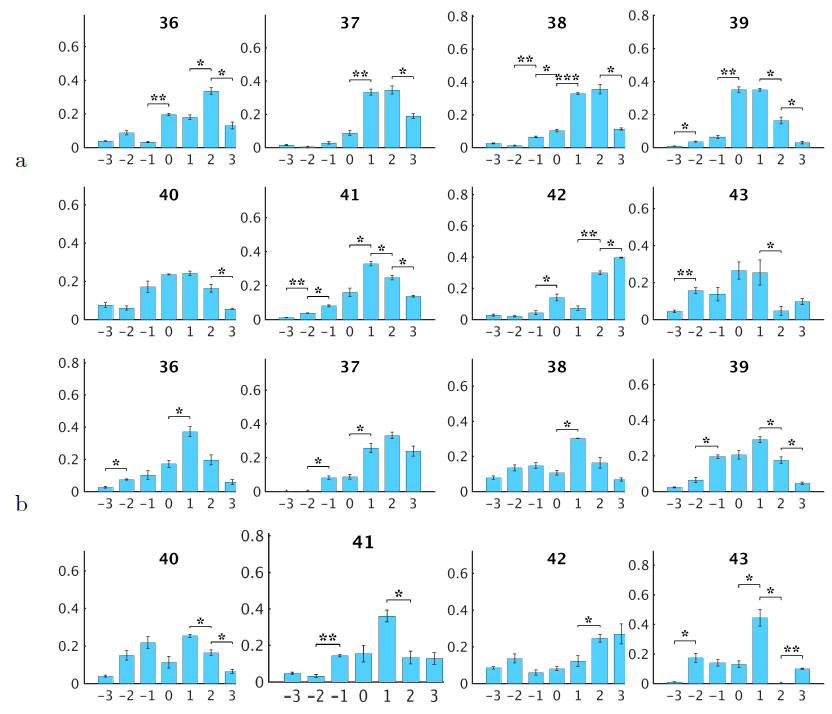}
\caption{\textbf{a.} Perception of arousal of the cultural-specific expression in cultures where these expressions are commonly used. Plots show the proportion of subjects' responses (y-axis) for the level of arousal indicated in the x-axis. Arousal is given by a 6-point scale, from less to more engaged. These results suggest a quite accurate perception of arousal for cultural-specific expressions, except for expression 36. Expression ID 37 has a mid to high arousal (values between 1 and 3 in the tested scale); expression ID 38 a value of 1 to 2; expressions IDs 39, 40 and 41 a value 1; expression ID 42 a value of 2 to 3; and expression ID 43 a value of 1. We note that all these values of arousal are on the positive side (i.e., the expression appears engaged). This may indicate a preference for emotive signals that convey alertness and active social interactions in cultural-specific expressions. \textbf{b.} Perception of arousal of the eight cultural-specific expressions in cultures where the expressions are not commonly observed. Note that although there is a tendency toward the results shown in a, there is little to no statistical significance between the top selection and those immediately right and left of it. The exception if expression ID 43, whose result is statistically significant and its perception of arousal matches that in \textbf{a}. * p \textless .01, ** p \textless .001, *** p \textless .001. Top-center number in each plot specifies the expression ID.}
\label{Fig: specific_arousal}
\end{figure}

We find that interpretation of arousal is relatively consistent, even if not highly accurate, Figure \ref{Fig: specific_arousal}a. Note the effect size of the most selected arousal value and the statistical significance between this and neighboring (left and right) levels of arousal. Additionally, note that the distribution in these plots is closer to that of an ideal observer (i.e., a Gaussian distribution with small variance). 

Given this latest result, it is worth asking whether arousal is also successfully interpreted by people in cultures where these expressions are not commonly observed. We find this not to be the case, Figure \ref{Fig: specific_arousal}b. 

In summary, the above results support a revised model of the production and perception of facial expressions of emotion with two types of expressions. The first type includes expressions that communicate emotion category as well as valence. These are used across cultures and, hence, have been referred herein as cross-cultural expressions. The second type of expressions do not transmit emotion categories but, rather, are best at communicating valence within and across cultures and arousal within cultures where those expressions are typically employed. We call these cultural-specific expressions. 

Of all affect concepts, valence is the one transmitted most robustly to an observer. Importantly, a linear machine learning classifier can correctly categorize (with 95.34\% accuracy) positive versus negative valence using just a few AUs, Section \ref{classifier_SDA}. 

Specifically, we find that positive valence is given by the presence of AU 6 and/or 12, whereas negative valence is given by the presence of AU 1, 4, 9 and/or 20. 

Hence, only six of the main 14 AUs are major contributors to the communication of valence. However, this result may only apply to expressions seen in day-to-day social interactions, and not where the intensity of valence is extreme since these are generally misinterpreted by observers \cite{aviezer2012body}. 

\subsubsection{Automatic classification of valence}\label{classifier_SDA} 

We define each facial expression as a 14-dimensional feature vector, with each dimension in this vector defining the presence (+1) or absence (0) of each AU. Sample vectors of the 43 facial expressions of emotion are grouped into two categories – those which yield the perception of positive valence and those yielding a perception of negative valence. 

Subclass Discriminant Analysis (SDA) \cite{zhu2006subclass} is used to identify the most discriminant AUs between these two categories – positive versus negative valence. This is given by the eigenvector, $\mathbf{v} = (v_1, \cdots v_{14})^T$, , associated with the largest eigenvalue, $\mathbf{\lambda} = (\lambda_1, \cdots \lambda_{14})^T$, of the metric of SDA. SDA's metric is computed as the product of between class dissimilarity and within class similarity. If the $i^{th}$ AU does not discriminate between the facial expressions of positive versus negative valence, then $\lambda \approx 0$. If $\lambda_{i} >> 0$, then this AU is present exclusively (or almost exclusively) in expressions of positive valence. If $\lambda_{i} << 0$, the $i^{th}$ AU is present exclusively (or almost exclusively) in expressions of negative valence. This process identified AUs 6 and 12 as major communicators of positive valence and AUs 1, 4, 9 and 20 as transmitters of negative valence. This result includes cross-cultural \textit{and} cultural-specific expressions.

We tested the accuracy of the above-defined SDA classifier using a leave-one-expression-out approach. That means the samples of all expressions, but one, are used to train the classifier and the sample defining the left-out expression is used to test it. With $n$ expressions, there are $n$ possible expressions that can be left out. We computed the mean classification accuracy of the $n$ possible left-out samples, given by the ratio of $\sum true positives + \sum true negatives$ over $\sum total population$. This yielded a classification accuracy of 95.34\%. That is, the AUs listed above are sufficient to correctly identify valence in a previously unseen expression 95.34\% of the time.\\

\noindent{\bf Data availability}: The data used in this paper is available from the authors and will be made accessible online (after acceptance). 

\section{Discussion}\label{Sec: Discussion}

To derive algorithms, researchers in affective computing use existing theoretical models of the production and visual perception of facial expressions of emotion. 

However, current theoretical models disagree on how many expressions are regularly used by humans and what affect variables these expressions transmit (emotion category, valence, arousal, a combination of them, or some other unknown, underlying dimensions) \cite{barrett2017emotions, russell2017science}. Thousands of experiments have been completed to test these models, with some supporting the transmission of emotion category and others favoring the transmission of valence, arousal and other dimensions \cite{martinez2017visual}. As a consequence, the area has been at an impasse for decades, with strong disagreements on how many expressions exist, what they communicate, and whether they are cultural-specific or cross-cultural \cite{chen2017discovering, ekman2016scientists, russell2003core, skerry2015neural, lindquist2016language}. 

We argue that previous studies were unable to resolve this impasse because they were performed in the laboratory, under highly controlled conditions. What is needed is an understanding of the expressions people of different cultures use in the wild, under unconstrained conditions. 

The goal is to discover these expressions, rather than search for pre-determined ones. The present paper constitutes, to our knowledge, the first of these studies. 

Specifically, we have studied 7.2 million facial configurations employed by people of 31 countries where either English, Spanish, Mandarin Chinese, Farsi, or Russian is the primary language, Table \ref{Table: common_label}. 

First, we found that while 16,384 facial configurations are possible, only 35 are typically observed across language groups. 

We then showed that these expressions successfully transmit emotion category and valence across cultures. 

These results are consistent with models of the production and perception of compound expressions \cite{du2014compound, martinez2017visual, barrett2010context} and models that suggest people experience at least 27 distinct emotions \cite{cowen2017self}. 

However, and crucially, the robustness of this categorical perception varies widely from expression to expression. This variability is not explained by current algorithms and model of the automatic analysis of the production and perception of facial expressions of emotion and, hence, they will need to be amended. 

One possible explanation for this variability in perceptual accuracy is that emotions are affected by constructs of people experiences \cite{barrett2017emotions, gendron2014cultural}. A related thesis is that this variability is explained by the number of neural mechanisms involved in the interpretation of facial expressions \cite{lindquist2012brain, spunt2017new, mather2016affective}. Even if that were the case though, we would still need to explain why some expressions are more affected than others. 

Moreover, it is worth noting that while the sensitivity of these categorizations is high, specificity is not, Table \ref{Table specificity}.  This suggests that the successful categorization of the herein identified expressions is important to humans, even if it comes at the cost of making errors elsewhere (i.e., false positives).

Importantly, we note that the number of facial configurations that communicate each emotion category varies significantly though, Table \ref{Table: common_label}. At one extreme, happiness is expressed using seventeen facial configurations. At the other end, disgust only utilizes a single expression. This result could point to individual differences in expressing the same emotion category \cite{du2014compound}, variations due to context \cite{etkin2015neural}, or multiple mechanisms as suggested by predictive coding \cite{barrett2017emotions}. But, at present, affective computing algorithms that analyze  the production and perception of facial expressions of emotion do not account for this effect. 

\begin{table}
\centering
\begin{tabular}{lll}
          & Specificity & Sensitivity \\ \hline \hline
Happiness & 0.616       & 0.961       \\
Sadness   & 0.487       & 0.967       \\
Anger     & 0.534       & 0.964       \\
Surprise  & 0.461       & 0.949       \\
Fear      & 0.418       & 0.967       \\
Disgust   & 0.349       & 0.983       \\ \hline \hline
\end{tabular}
\caption{Specificity and sensitivity of the visual recognition of the six basic emotion categories percieved by participants} \label{Table specificity}
\end{table}

Second, we discovered that the number of expressions that are used in one or more, but not all, cultures is even smaller than that of cross-cultural expressions. The number of these cultural-specific expressions is 8. 

This result suggests that most facial expressions of emotion are used across cultures, as some theories of emotion propound \cite{ekman1992argument, izard2013human}. Crucially though, a subsequent study of the visual perception of cultural-specific expressions showed they successfully convey valence and arousal, but not emotion category. More striking was the finding that valence, but not arousal, is effectively interpreted in cultures where these cultural-specific expressions are not regularly employed. This is another finding that cannot be explained by current models and, consequently, is not coded in current affective computing algorithms.  

Some studies in cognitive neurosceince have suggested there are dissociated brain mechanisms for the recognition of emotion categories versus valence versus arousal, \cite{chikazoe2014population, harris2012morphing, kim2017human, srinivasan2016neural, wilson2013neural}. These results are consistent with our finding of two sets of expressions, with only one set transmitting emotion category but both successfully communicating valence. Additionally, only the cultural-specific expressions convey arousal, and, even here, only within the cultures where those expressions are commonly observed. This is consistent with the work of \cite{kensinger2004two} who found that distinct neural processes are involved in emotional memory formation for arousing versus valenced, non-arousing stimuli. However, our results seem to contradict those of a recent study showing robust transmission of arousal across cultures and species \cite{filippi2017humans}. Nonetheless, it is worth noting that this latter study involved the recognition of arousal of an aural signal, not a visual one. This could suggest that arousal is most robustly transmitted orally. 

Adding to this differentiation between oral and visual emotive signals \cite{sauter2010cross} found that negative valence is successfully communicated orally across cultures, but that most culture-specific oral signals have positive valence. In contrast to these findings, our results suggest that the visual signal of emotion conveys positive and negative valence across cultures, and mostly (but not exclusively) positive valence in culture-specific expressions. Thus, there appears to be a more extensive communication of valence by the visual signal than by the oral one. 

The above results are in support of a dedicated neural mechanism for the recognition of valence. A good candidate for this neural mechanism is the amygdala \cite{wang2017human}. For instance, \cite{kim2017human} have recently shown that the amygdala differentially responds to positive versus negative valence even when the emotion category is constant; although the amygdala may also determine arousal \cite{wilson2013neural}. Another candidate is the subthalamic nucleus. Using single-cell recordings in humans, \cite{kim2017human} identified dissociated cell populations in this brain regions for the recognition of arousal and valence. But, it is also likely that multiple areas contribute to the recognition of valence as well as other affect signals \cite{lindquist2012brain, mather2016affective, spunt2017new}.

Thus, the cognitive neuroscience studies cited in the preceding paragraphs are in support of our results and, hence, affective computing methods must include them if the results are to be meaningful to users. We thus posit that the results presented herein will be fundamental to advance the state of the art in the area. New or amended models of emotion will need to be proposed.

\section{Conclusions}\label{Sec: Conclusions}

Emotions play a major role in human behavior. Affective computing is attempting to infer as much information as possible from the aural and visual signals.

Yet, at present, researchers disagree on how many emotions humans have, whether these are cross-cultural or cultural-specific, whether emotions are best represented categorically or as a set of continuous affect variables (valence, arousal), how context influences production and perception of expressions of emotion, or whether personal constructs modulate the production and perception of emotion. 

Laboratory studies and controlled in-lab experiments are insufficient to address this impasse. What is needed is a comprehensive study of which expressions exist in the wild (i.e., in uncontrolled conditions outside the laboratory), not in tailored, controlled experiments. 

To address this problem, the present paper presented the first large-scale study (7.2 million images) of the production and visual perception of facial expressions of emotion in the wild. We found that of the 16,384 possible facial configurations that people can produce only 35 are successfully used to transmit affect information across cultures, and only 8 within a smaller number of cultures. 

Crucially, our studies showed that these 35 cross-cultural expressions successfully transmit emotion category and valence but not arousal. The 8 cultural-specific expressions successfully transmit valence and, to some degree, arousal, but not emotion category. 

We also find that the degree of successful visual interpretation of these 43 expressions varies significantly. And, the number of expressions used to communicate each emotion is also different, e.g., 17 expressions transmit happiness, but only 1 is used to convey disgust. 

These  findings are essential to change the direction in the design of algorithms for the coding and reading of emotion.

\section*{Acknowledgment}
This research was supported by the National Institutes of Health (NIH), grant R01-DC-014498, and the Human Frontier Science Program (HFSP), grant RGP0036/2016.

  %

\ifCLASSOPTIONcaptionsoff
  \newpages
\fi



%

\bibliography{bare_jrnl_compsoc}
\bibliographystyle{IEEEtran}

\end{document}